\documentclass[runningheads]{llncs}

\usepackage[T1]{fontenc}

\usepackage{graphicx}
\usepackage{listings}
\usepackage{cite}
\usepackage[colorlinks=true, allcolors=blue]{hyperref}
\usepackage{multirow}
\usepackage{amsmath}
\usepackage{amssymb}
\usepackage{mathtools}

\newcommand\Tstrut{\rule{0pt}{2.5ex}}         
\newcommand\Bstrut{\rule[-0.9ex]{0pt}{0pt}}   

\begin{document}
\title{Joint Denoising and Few-angle Reconstruction for Low-dose Cardiac SPECT Using a Dual-domain Iterative Network with Adaptive Data Consistency}
\titlerunning{Submission 387}


\author{Xiongchao Chen \and
Bo Zhou \and
Huidong Xie \and
Xueqi Guo \and
Qiong Liu \and
Albert J. Sinusas \and
Chi Liu}
\authorrunning{X. Chen et al.}

\institute{Yale University, New Haven, CT 06511, USA \\
\email{\{xiongchao.chen, chi.liu\}@yale.edu}}

\maketitle     

\begin{abstract}
Myocardial perfusion imaging (MPI) by single-photon emission computed tomography (SPECT) is widely applied for the diagnosis of cardiovascular diseases. Reducing the dose of the injected tracer is essential for lowering the patient's radiation exposure, but it will lead to increased image noise. Additionally, the latest dedicated cardiac SPECT scanners typically acquire projections in fewer angles using fewer detectors to reduce hardware expenses, potentially resulting in lower reconstruction accuracy. To overcome these challenges, we propose a dual-domain iterative network for end-to-end joint denoising and reconstruction from low-dose and few-angle projections of cardiac SPECT. The image-domain network provides a prior estimate for the projection-domain networks. The projection-domain primary and auxiliary modules are interconnected for progressive denoising and few-angle reconstruction. Adaptive Data Consistency (ADC) modules improve prediction accuracy by efficiently fusing the outputs of the primary and auxiliary modules. Experiments using clinical MPI data show that our proposed method outperforms existing image-, projection-, and dual-domain techniques, producing more accurate projections and reconstructions. Ablation studies confirm the significance of the image-domain prior estimate and ADC modules in enhancing network performance. The source code is released at \href{https://***.com}{https://***.com}.

\end{abstract}

\section{Introduction}
Myocardial perfusion imaging (MPI) through Single-Photon Emission Computed Tomography (SPECT) is the most commonly employed exam for diagnosing cardiovascular diseases \cite{danad2017comparison, gimelli2009stress, nishimura2008prognostic}. However, exposure to ionizing radiation from SPECT radioactive tracers presents potential risks to both patient and healthcare provider \cite{einstein2012effects}. While reducing the injected dose can lower the radiation exposure, it will lead to increased image noise \cite{henzlova2016asnc}. Additionally, acquiring projections in fewer angles using fewer detectors is a viable strategy for shortening scanning time and reducing hardware expenses. However, fewer-angle projections can lead to lower reconstruction accuracy and higher image noise \cite{niu2014sparse, zhu2013improved}.


Many deep learning methods by Convolutional Neural Networks (CNNs) have been developed for denoising or few-angle reconstruction in nuclear medicine. Existing techniques for denoising in nuclear medicine were implemented either in the projection or image domain. Low-dose (LD) projection or image was input to CNN to predict the corresponding full-dose (FD) projection \cite{sun2022deep, aghakhan2022deep} or image \cite{ramon2020improving, sun2022pix2pix, liu2021pet, wang20183d}. Previous approaches for few-angle reconstruction in nuclear medicine were developed based on projection-, image-, or dual-domain frameworks. In the projection- or image-domain methods, the few-angle projection or image was input to CNN to generate the full-angle projection \cite{whiteley2019cnn, shiri2020standard} or image \cite{amirrashedi2021deep}, respectively. The dual-domain method, Dual-domain Sinogram Synthesis (DuDoSS), utilized the image-domain output as an initial estimation for the prediction of the full-angle projection in the projection domain \cite{chen2022dudoss}.

The latest dedicated cardiac SPECT scanners tend to employ fewer detectors to minimize hardware costs \cite{wu2019recent, gehealthcaremyospectessystems}. Although deep learning-enabled denoising or few-angle reconstruction in nuclear medicine has been extensively studied in previous works, end-to-end joint denoising and few-angle reconstruction for the latest dedicated scanners still remains highly under-explored. Here, we present a dual-domain iterative network with learnable Adaptive Data Consistency (ADC) modules for joint denoising and few-angle reconstruction of cardiac SPECT. The image-domain network provides a prior estimate for the prediction in the projection domain. Paired primary and auxiliary modules are interconnected for progressive denoising and few-angle restoration. ADC modules are incorporated to enhance prediction accuracy by fusing the predicted projections from primary and auxiliary modules. We evaluated the proposed method using clinical data and compared it to existing projection-, image-, and dual-domain methods. In addition, we also conducted ablation studies to assess the impact of the image-domain prior estimate and ADC modules on the network performance.

\section{Methods}
\subsection{Dataset and Pre-processing}
A dataset consisting of 474 anonymized clinical hybrid SPECT-CT MPI studies was collected. Each study was conducted following the injection of $^{99\mathrm{m}}$Tc-tetrofosmin on a GE NM/CT 570c \cite{wu2019recent}. The clinical characteristics of enrolled patients are listed in supplementary Table S1. 

\begin{figure}[htb!]
\centering
\includegraphics[width=0.99\textwidth]{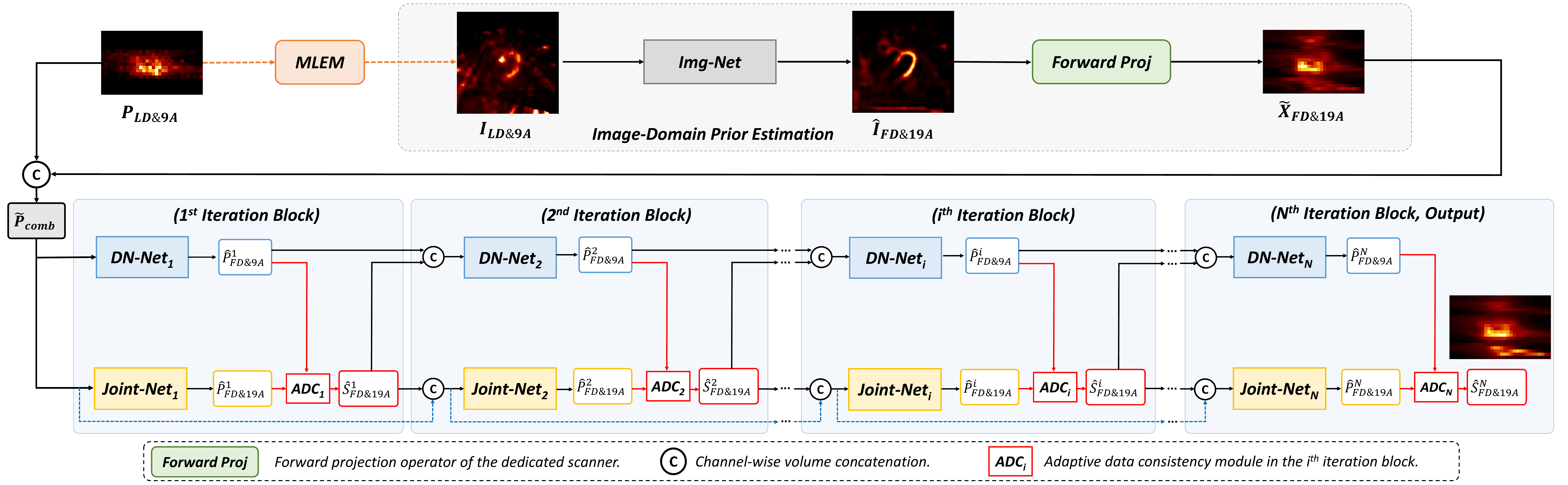}
\caption{Dual-domain iterative network. The forward projection of the image-domain output (top row) serves as a prior estimate for the projection domain (bottom row). The primary ($\textit{Joint-Net}$) and the auxiliary modules ($\textit{DN-Net}$) are interconnected in every iteration for progressive denoising and few-angle restoration.}
\label{fig:overview}
\end{figure}

The GE 530c/570c scanners comprise of 19 pinhole detectors arranged in three columns on a cylindrical surface \cite{chan2016impact}. The few-angle projections were generated by selecting the 9 angles (9A) in the central column, simulating the configurations of the latest cost-effective MyoSPECT ES few-angle scanner \cite{gehealthcaremyospectessystems} as shown in supplementary Fig. S1. The 10$\%$-dose LD projections were produced by randomly decimating the list-mode data at a 10$\%$ downsampling rate. The simulated LD$\&$9A projection is the input, and the original FD$\&$19A projection is the label. We used 200, 74, and 200 cases for training, validation, and testing.

\subsection{Dual-domain Iterative Network}
The dual-domain iterative network is shown in Fig.~\ref{fig:overview}. The LD$\&$9A projection $P_{LD\&9A}$ is first input into a Maximum-Likelihood Expectation Maximization (MLEM, 30 iterations) module, reconstructing the LD$\&$9A image $I_{LD\&9A}$. \\

\noindent \textbf{Image-Domain Prior Estimation.} $I_{LD\&9A}$ is then input to the image-domain network \textit{Img-Net}, i.e. a CNN module, to produce the predicted image $\hat{I}_{FD\&19A}$, supervised by the ground-truth FD$\&$19A image $I_{FD\&19A}$. The image-domain loss $\mathcal{L}_I$ is:
\begin{equation}
     \mathcal{L}_I = \left\|\hat{I}_{FD\&19A} - I_{FD\&19A}\right\|_1.
\end{equation}    

\noindent Then, $\hat{I}_{FD\&19A}$ is fed into a forward projection (FP) operator of GE 530c/570c, producing $\widetilde{X}_{FD\&19A}$ as the prior estimate of the ground-truth FD$\&$19A projection $P_{FD\&19A}$. The image-domain prediction can be formulated as:
\begin{equation}
    \widetilde{X}_{FD\&19A} = \mathcal{F}({\mathcal{I}(I_{LD\&9A})}),
\end{equation}
where $\mathcal{I}(\cdot)$ is the \textit{Img-Net} operator and $\mathcal{F(\cdot)}$ is the FP operator. \\

\noindent \textbf{Projection-Domain Iterative Prediction.} The prior estimate $\widetilde{X}_{FD\&19A}$ is then channel-wise concatenated with $P_{LD\&9A}$ to generate $\widetilde{P}_{comb}$, which serves as the input to the projection-domain networks, formulated as:
\begin{equation}
   \widetilde{P}_{comb} = \left\{\widetilde{X}_{FD\&19A}, P_{LD\&9A}\right\},
\end{equation}
where $\left\{ \cdot \right\}$ refers to channel-wise concatenation of 3D projections. 

Given the difficulty in performing joint denoising and few-angle restoration directly, we split the two tasks and assign them to two parallel Attention U-Net modules \cite{ronneberger2015u}: the auxiliary module for denoising ($\textit{DN-Net}$) and the primary module for joint prediction ($\textit{Joint-Net}$). In each iteration, $\textit{DN-Net}$ solely focuses on denoising and produces an auxiliary projection. $\textit{Joint-Net}$ performs both denoising and few-angle restoration, producing the primary projection. The auxiliary and primary projections are then fused in an ADC module (described in subsection 2.3), producing a fused projection of higher accuracy.

\begin{figure}[htb!]
\centering
\includegraphics[width=0.85\textwidth]{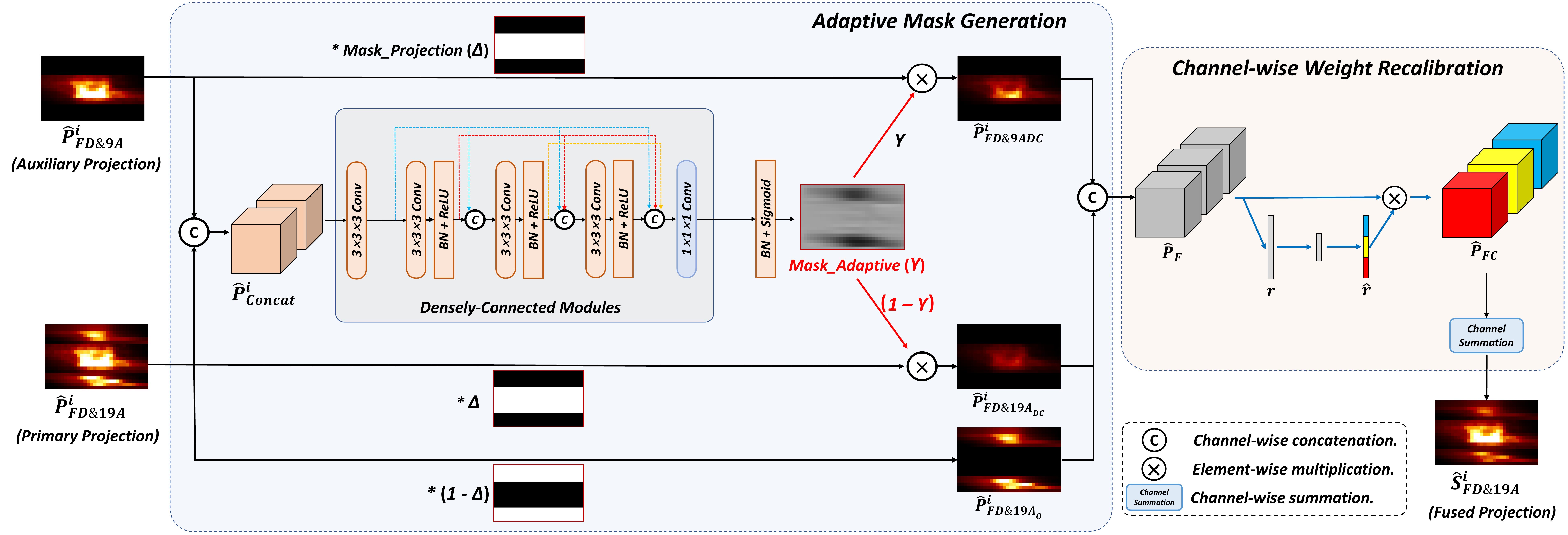}
\caption{Adaptive data consistency (ADC) is composed of an Adaptive Mask Generation module (left) to fuse the auxiliary and the primary projections and a Channel-wise Weight Recalibration module (right) to adjust weights of the combined projections.}
\label{fig:adc}
\end{figure}

In the 1$^\mathrm{st}$ iteration block, $\widetilde{P}_{comb}$ is input to $\textit{DN-Net}_1$ to produce the auxiliary projection $\hat{P}_{FD\&9A}^1$. It is also input to $\textit{Joint-Net}_1$ to produce the primary projection $\hat{P}_{FD\&19A}^1$. Then, $\hat{P}_{FD\&9A}^1$ and $\hat{P}_{FD\&19A}^1$ are fused in the $\textit{ADC}_1$ module, producing the fused projection $\hat{S}_{FD\&19A}^1$, formulated as:
\begin{equation}
    \hat{S}_{FD\&19A}^1 = \mathcal{A}_1(\mathcal{D}_1(\widetilde{P}_{comb}), \mathcal{J}_1(\widetilde{P}_{comb})),
\end{equation}
where $\mathcal{A}_1(\cdot)$ is the $\textit{ADC}_1$ operator. $\mathcal{D}_1(\cdot)$ is the $\textit{DN-Net}_1$ operator, and $\mathcal{J}_1(\cdot)$ is the $\textit{Joint-Net}_1$ operator.

In the $i^{th} (i\geq2)$ iteration, the output of the ${(i-1)}^{th}$ iteration, $\hat{S}_{FD\&19A}^{(i-1)}$, was added to the input of 
$\textit{DN-Net}_{i}$ to assist the denoising of the auxiliary module. $\hat{S}_{FD\&19A}^{(i-1)}$ is concatenated with the output of $\textit{DN-Net}_{(i-1)}$ and then fed into $\textit{DN-Net}_i$ to produce the auxiliary projection $\hat{P}_{FD\&9A}^i$ in the $i_{th}$ iteration:
\begin{equation}
    \hat{P}_{FD\&9A}^i = \mathcal{D}_{i}(\left\{\hat{S}_{FD\&19A}^{(i-1)}, \hat{P}_{FD\&9A}^{(i-1)} \right\}),
\end{equation}
where $\mathcal{D}_i(\cdot)$ is the $\textit{DN-Net}_i$ operator. Then, the outputs of all the $(i-1)$ previous iterations, $\hat{S}_{FD\&19A}^{m} (m<i)$, are densely connected with $\widetilde{P}_{comb}$ as the input to $\textit{Joint-Net}_i$ to produce the primary projection in the $i_{th}$ iteration:
\begin{equation}
    \hat{P}_{FD\&19A}^i = \mathcal{J}_{i}(\left\{\widetilde{P}_{comb}, \hat{S}_{FD\&19A}^{1}, \hat{S}_{FD\&19A}^{2}, \cdots, \  \hat{S}_{FD\&19A}^{(i-1)}\right\}),
\end{equation}
where $\mathcal{J}_i(\cdot)$ is the $\textit{Joint-Net}_i$ operator. Then, the auxiliary and primary projections are fused in $\textit{ADC}_i$ for recalibration, generating the fused $\hat{S}_{FD\&19A}^{i}$ as:
\begin{equation}
    \hat{S}_{FD\&19A}^{i} = \mathcal{A}_{i}(\hat{P}_{FD\&9A}^i, \hat{P}_{FD\&19A}^i),
\end{equation}
where $\mathcal{A}_i(\cdot)$ is the $\textit{ADC}_i$ operator. The overall network output $\hat{S}_{FD\&19A}^{N}$ is the output of the $N^{\mathrm{th}}$ iteration, where $N$ is the total number of iterations with a default value of 4. The projection-domain loss is formulated as:
\begin{equation}
    \mathcal{L}_P = \sum_{i=1}^{N}(\left\| \hat{P}_{FD\&9A}^{i} - P_{FD\&9A} \right\|_1 + \left\| \hat{S}_{FD\&19A}^{i} - P_{FD\&19A} \right\|_1),
\end{equation}
where $P_{FD\&9A}$ is the FD$\&$9A projection. The total loss function $\mathcal{L}$ is the weighted summation of the image-domain loss $\mathcal{L}_I$ and the projection-domain loss $\mathcal{L}_P$: 
\begin{equation}
    \mathcal{L} = w_I \mathcal{L}_I + w_P \mathcal{L}_P,
\end{equation}
where the weights $w_I$ and $w_P$ were empirically set as 0.5 in our experiment.

\begin{figure}[htb!]
\centering
\includegraphics[width=0.90\textwidth]{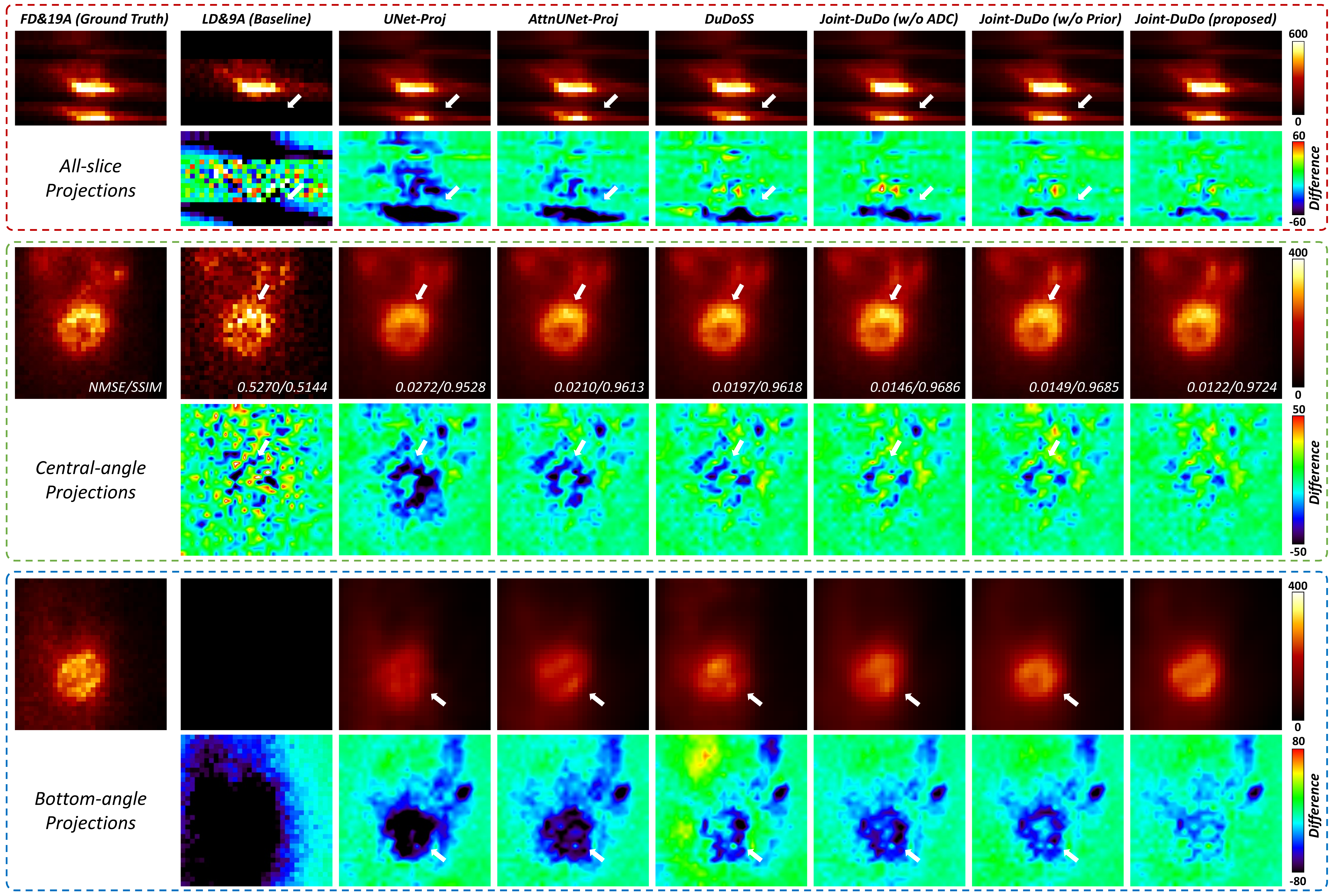}
\caption{Predicted FD$\&$19A projections in side, central-angle, and bottom-angle views with NMSE/SSIM annotated. White arrows denote prediction inconsistency.}
\label{fig:proj}
\end{figure}

\subsection{Adaptive Data Consistency}
The initial data consistency (DC) was used to fuse the predicted and ground-truth k-spaces, thereby ensuring the consistency of the MRI image in k-space \cite{chlemper2017deep, qin2018convolutional,zhou2020dudornet}. It is also utilized in the DuDoSS for the few-angle reconstruction of cardiac SPECT imaging \cite{chen2022dudoss}. However, in our study, the ground-truth FD$\&$9A projection $P_{FD\&9A}$, which is a pre-requisite for applying DC to $\hat{P}_{FD\&19A}$, is not available as input. Thus, we generate $\hat{P}_{FD\&9A}^{i}$ using $\textit{DN-Net}_i$ as the intermediate auxiliary information to improve $\hat{P}_{FD\&19A}$. The proposed ADC generates a voxel-wise adaptive projection mask for the fusion of $\hat{P}_{FD\&9A}^{i}$ and $\hat{P}_{FD\&19A}^{i}$. 

As presented in Fig.~\ref{fig:adc}, in the $i^{th}$ iteration, $\hat{P}_{FD\&9A}^{i}$ and $\hat{P}_{FD\&19A}^{i}$ are first concatenated and input to a densely-connected \cite{huang2017densely} CNN module for spatial feature extraction. Then, a voxel-wise adaptive projection mask $\gamma$ is generated from the extracted features using a Sigmoid operator, which determines the voxel-wise weights (from 0 to 1) for the summation of $\hat{P}_{FD\&9A}^{i}$ and $\hat{P}_{FD\&19A}^{i}$. The weighted projections of the central columns are generated as:
\begin{equation}
    \hat{P}_{FD\&9A_{DC}}^{i}  = \hat{P}_{FD\&9A}^{i}   \ast  \Delta  \ast \gamma,  
\end{equation}
\begin{equation}
    \hat{P}_{FD\&19A_{DC}}^{i} = \hat{P}_{FD\&19A}^{i}  \ast  \Delta  \ast (1 - \gamma),
\end{equation}
where $\ast$ is the voxel-wise multiplication, and $\Delta$ refers to the binary mask of the few-angle projection (shown in Fig.~\ref{fig:adc}). In addition, the outer columns of $\hat{P}_{FD\&19A}$ is computed as: $\hat{P}_{FD\&19A_O}^{i} = \hat{P}_{FD\&19A}^{i}  \ast  (1 - \Delta)$.

Then, the above three weighted projections are concatenated and input to a Channel-wise Weight Recalibration module, a squeeze-excitation \cite{hu2018squeeze} self-attention mechanism, to generate a channel recalibration vector $\hat{r} = [r_1, r_2, r_3]$. The output of ADC is the weighted summation of the recalibrated projections as:
\begin{equation}
    \hat{S}_{FD\&19A}^{i} = r_1 \hat{P}_{FD\&9A_{DC}}^{i} + r_2 \hat{P}_{FD\&19A_{DC}}^{i} + r_3 \hat{P}_{FD\&19A_{O}}^{i}.
\end{equation}

\subsection{Implementation Details}
We evaluated Joint-DuDo against various deep learning methods in this study. Projection-domain methods using U-Net (designated as UNet-Proj) \cite{ronneberger2015u} or Attention U-Net (designated as AttnUNet-Proj) \cite{oktay2018attention}, the image-domain method using Attention U-Net (designated as AttnUNet-Img) \cite{whiteley2019cnn}, and the dual-domain method DuDoSS \cite{chen2022dudoss} were tested. We also included ablation study groups without ADC (but with normal DC, designated as Joint-DuDo (w/o ADC)) or without the image-domain prior estimate (designated as Joint-DuDo (w/o Prior)).

Networks were developed using PyTorch \cite{paszke2019pytorch} and trained with Adam optimizers \cite{kingma2014adam}. The initial learning rate was $10^{-3}$ for image and projection modules and $10^{-4}$ for ADC modules, with a decay rate of 0.99 per epoch to avoid overfitting \cite{you2019does}. Joint-DuDo and ablation groups were trained for 50 epochs and the other groups were trained for 200 epochs. The default number of iterations of Joint-DuDo was 4. Evaluations of Joint-DuDo using multiple iterations (1 to 6) are shown in section 3 (Fig.~\ref{fig:plot} left). Evaluations of more datasets with different LD levels (1 to 80$\%$, default 10$\%$) are shown in section 3 (Fig.~\ref{fig:plot} mid, right).

\begin{table} [htb!]
\caption{Evaluation of the predicted FD$\&$19A projections using normalized mean square error (NMSE), normalized mean absolute error (NMAE), structural similarity (SSIM), and peak signal-to-noise ratio (PSNR). The best results are in \textcolor{red}{red}.}
\label{tab:proj} 
\tiny
\centering
\begin{tabular}{ l | c | c | c | c || c}

\hline
\textbf{Methods}         & \textbf{NMSE($\boldsymbol{\%}$)}     & \textbf{NMAE($\boldsymbol{\%}$)}    & \textbf{SSIM}                   & \textbf{PSNR}                & \textbf{P-values$^{\dag}$}  \Tstrut\Bstrut\\  
\hline 
Baseline LD-9A           & $54.46 \pm 2.46$                     & $62.44 \pm 2.53$                    & $0.4912 \pm 0.0260$             & $19.23 \pm 1.68$             & < 0.001                      \Tstrut\Bstrut\\
\hline
UNet-Proj  \cite{ronneberger2015u}         & $4.26 \pm 1.58$                      & $16.65 \pm 2.42$                    & $0.9247 \pm 0.0248$             & $30.54 \pm 1.79$             & < 0.001                      \Tstrut\Bstrut\\  
AttnUNet-Proj \cite{oktay2018attention}             & $3.43 \pm 1.17$                      & $15.34 \pm 2.40$                    & $0.9372 \pm 0.0193$             & $31.47 \pm 1.65$             & < 0.001                      \Tstrut\Bstrut\\  
DuDoSS \cite{chen2022dudoss}              & $3.10 \pm 0.78$                      & $14.54 \pm 1.59$                    & $0.9429 \pm 0.0153$             & $31.82 \pm 1.50$             & < 0.001                      \Tstrut\Bstrut\\  
Joint-DuDo (w/o ADC)    & $2.42 \pm 0.81$                      & $13.04 \pm 1.89$                    & $0.9509 \pm 0.0156$             & $32.97 \pm 1.60$             & < 0.001                      \Tstrut\Bstrut\\  
Joint-DuDo (w/o Prior)  & $2.43 \pm 0.82$                      & $13.36 \pm 1.90$                    & $0.9507 \pm 0.0150$             & $32.95 \pm 1.68$             & < 0.001                      \Tstrut\Bstrut\\  
Joint-DuDo (proposed)            & \textcolor{red}{$2.16 \pm 0.71$}             & \textcolor{red}{$12.51 \pm 1.73$}           & \textcolor{red}{$0.9548 \pm 0.0137$}    & \textcolor{red}{$33.47 \pm 1.67$}    & \textendash                 \Tstrut\Bstrut\\  
\hline

\multicolumn{6}{l}{$^{\dag}$P-values of the paired t-tests of NMSE between the current method and Joint-DuDo (proposed).} \\
\end{tabular}
\end{table}

\section{Results}
Fig.~\ref{fig:proj} presents the qualitative comparison of the predicted FD$\&$19A projections in different projection views. It can be observed that Joint-DuDo generates more accurate predicted projections at all views compared to the projection- and dual-domain approaches. Joint-DuDo also demonstrates higher accuracy compared to the ablation study groups without the image-domain prior estimate or ADC modules, proving the roles of the prior estimate and ADC in enhancing network performance. Table~\ref{tab:proj} outlines the quantitative evaluations of the predicted projections. Joint-DuDo outperforms existing projection- and dual-domain approaches and the ablation study groups ($p<0.001$).

\begin{figure}[htb!]
\centering
\includegraphics[width=0.88\textwidth]{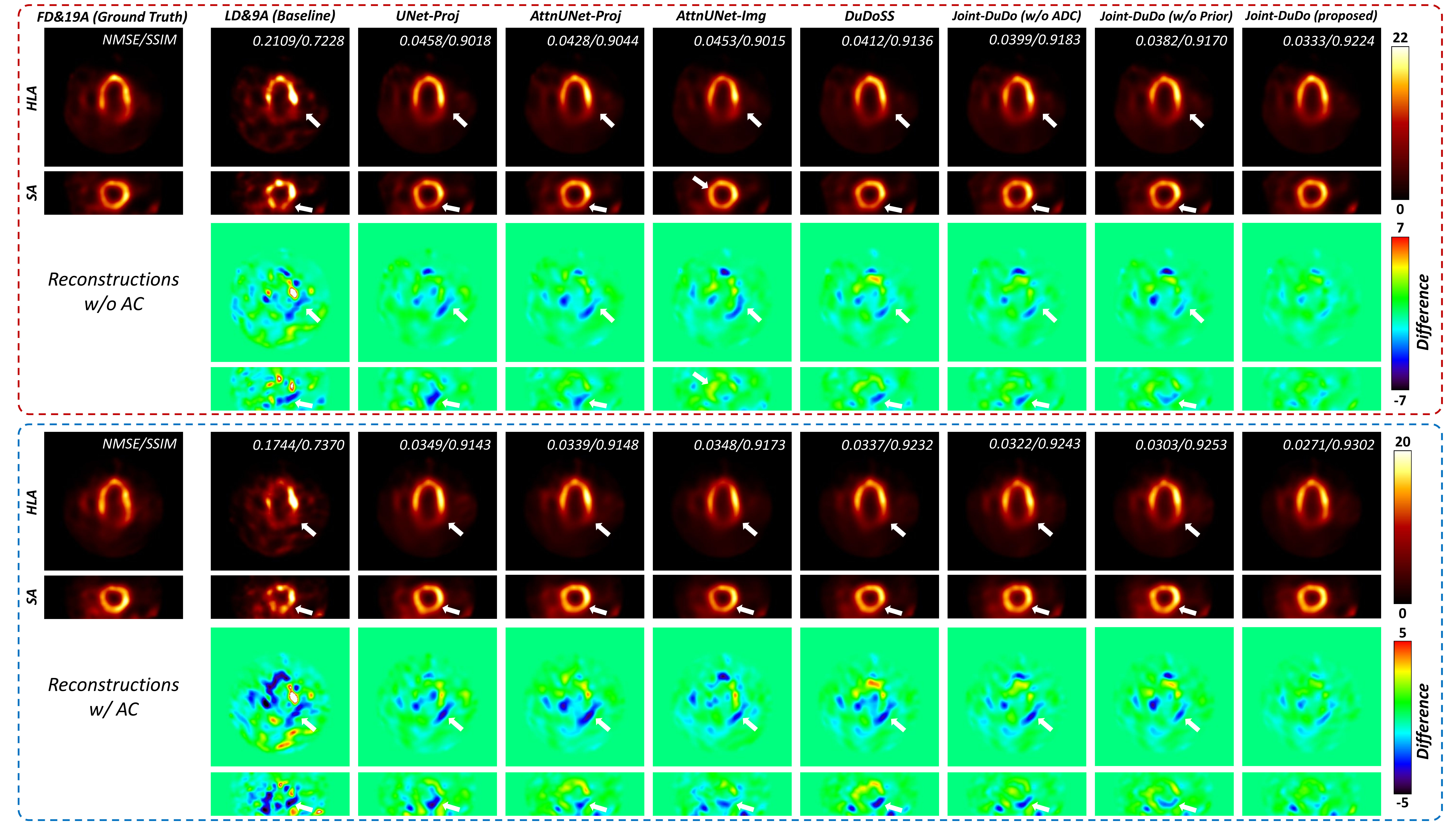}
\caption{Reconstructed or predicted FD$\&$19A SPECT images w/ or w/o the CT-based attenuation correction (AC). White arrows denote the prediction inconsistency.}
\label{fig:recon}
\end{figure}

\begin{table} [htb!]
\caption{Evaluation of the reconstructed or predicted FD$\&$19A SPECT images w/ or w/o attenuation correction (AC). The best results are marked in \textcolor{red}{red}.}
\label{tab:recon} 
\tiny
\centering
\begin{tabular}{ l | c | c | c | c | c | c }

\hline
\multirow{2}{*}{\textbf{Methods}}      & \multicolumn{3}{|c|}{\textbf{Reconstructed Images w/o AC}}     & \multicolumn{3}{|c}{\textbf{Reconstructed Images w/ AC}}                        \Tstrut\Bstrut\\
\cline{2-7}
                        & \textbf{NMSE($\boldsymbol{\%}$)}   & \textbf{NMAE($\boldsymbol{\%}$)}   & \textbf{PSNR}  & \textbf{NMSE($\boldsymbol{\%}$)}      & \textbf{NMAE($\boldsymbol{\%}$)}  & \textbf{PSNR}  \Tstrut\Bstrut\\
\hline 
Baseline LD-9A           & $30.66 \pm 11.27$             & $47.76 \pm 6.74$             & $25.41 \pm 2.12$             & $22.57 \pm 9.42$            & $42.23 \pm 7.02$            & $26.17 \pm 2.05$          \Tstrut\Bstrut\\
\hline 
UNet-Proj \cite{ronneberger2015u}     & $7.00 \pm 2.56$               & $23.59 \pm 3.75$             & $31.77 \pm 1.94$             & $5.39 \pm 2.02$             & $20.85 \pm 3.34$            & $32.26 \pm 1.75$          \Tstrut\Bstrut\\  
AttnUNet-Proj \cite{oktay2018attention}   & $6.14 \pm 2.08$               & $22.32 \pm 3.29$             & $32.33 \pm 1.96$             & $4.70 \pm 1.65$             & $19.61 \pm 2.94$            & $32.85 \pm 1.76$          \Tstrut\Bstrut\\ 
AttnUNet-Img \cite{whiteley2019cnn}          & $6.07 \pm 1.44$               & $21.80 \pm 2.26$             & $32.29 \pm 1.87$             & $4.66 \pm 1.05$             & $19.50 \pm 1.95$            & $32.78 \pm 1.59$          \Tstrut\Bstrut\\ 
DuDoSS \cite{chen2022dudoss}     & $5.57 \pm 1.77$               & $21.70 \pm 3.11$             & $32.61 \pm 1.93$             & $4.44 \pm 1.40$             & $18.89 \pm 2.84$            & $33.09 \pm 1.75$          \Tstrut\Bstrut\\   
Joint-DuDo (w/o ADC)    & $5.32 \pm 1.78$               & $20.57 \pm 3.12$             & $32.95 \pm 1.91$             & $4.22 \pm 1.44$             & $18.42 \pm 2.83$            & $33.32 \pm 1.73$          \Tstrut\Bstrut\\  
Joint-DuDo (w/o Prior)  & $5.45 \pm 1.83$               & $20.75 \pm 3.20$             & $32.84 \pm 1.93$             & $4.34 \pm 1.51$             & $18.60 \pm 2.85$            & $33.20 \pm 1.76$          \Tstrut\Bstrut\\  
Joint-DuDo (proposed)   & \textcolor{red}{$4.68 \pm 1.46$}      & \textcolor{red}{$19.27 \pm 2.70$}    & \textcolor{red}{$33.49 \pm 1.87$}    & \textcolor{red}{$3.72 \pm 1.19$}    & \textcolor{red}{$17.32 \pm 2.48$}   & \textcolor{red}{$33.85 \pm 1.71$}  \Tstrut\Bstrut\\
\hline 

\end{tabular}
\end{table}

Fig.~\ref{fig:recon} shows the qualitative comparison of the reconstructed or predicted FD$\&$19A images with or without the CT-based attenuation correction (AC). Joint-DuDo results in more accurate SPECT images compared to other image-domain, projection-domain, and dual-domain approaches as well as the ablation groups. Segment-wise visualizations of the images in Fig.~\ref{fig:recon} are shown in supplementary Fig. S2 and S3. With or without AC, Joint-DuDo outperforms the other methods ($p<0.001$) as indicated by the quantitative comparison in Table~\ref{tab:recon}.

\begin{figure}[htb!]
\centering
\includegraphics[width=0.90\textwidth]{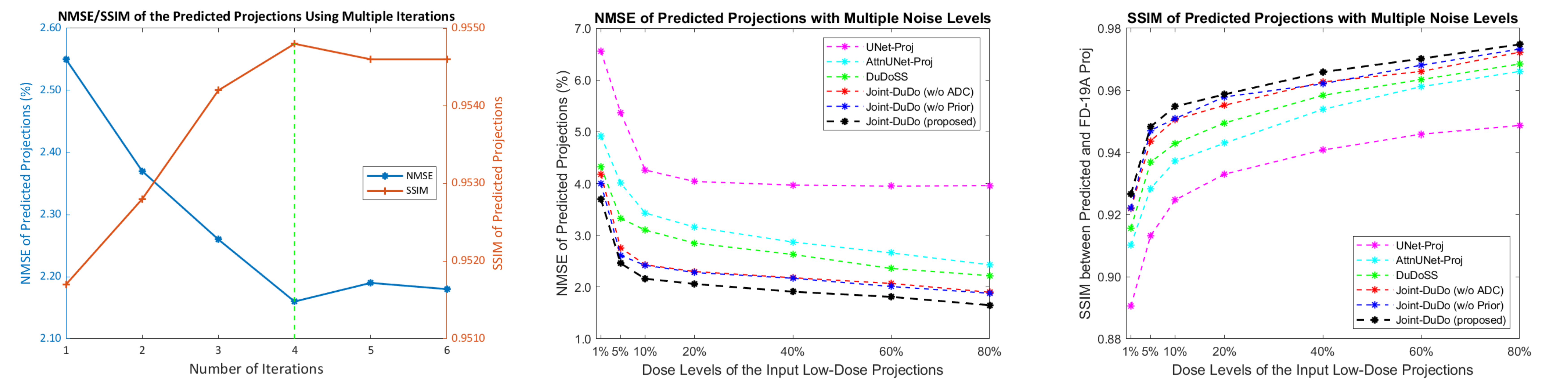}
\caption{Evaluation of Joint-DuDo using multiple iterations (left). Evaluation of various approaches based on datasets with different LD levels (mid, right).}
\label{fig:plot}
\end{figure}

As shown in Fig.~\ref{fig:plot} (left), the performance of Joint-DuDo improves as the number of iterations ($N$) increases and reaches convergence at $N=4$. In addition, we generated more datasets with different LD levels (1$\%$ to 80$\%$) to test the network performance as shown in Fig.~\ref{fig:plot} (mid and right). It can be observed that our proposed Joint-DuDo demonstrates consistently higher prediction accuracy across various LD levels compared to other testing methods.

\section{Discussion and Conclusion}
In this work, we propose Joint-DuDo, a novel dual-domain iterative network with learnable ADC modules, for the joint denoising and few-angle reconstruction of low-dose cardiac SPECT. Joint-DuDo employs the output of the image domain as an initial estimate for the projection prediction in the projection domain. This initial estimate enables the input closer to the target, thus enhancing the overall prediction accuracy. The ADC modules produce adaptive projection masks to fuse the predicted auxiliary and primary projections for higher output accuracy. Experiments using clinical data showed that the proposed Joint-DuDo led to higher accuracy in the projections and reconstructions than existing projection-, image-, and dual-domain approaches. 

The potential clinical significance of our work is that it shows the feasibility of simultaneously performing denoising and few-angle reconstruction in low-dose cardiac SPECT. Using the proposed method, we could potentially promote the clinical adoption and market coverage of the latest cost-effective fewer-angle SPECT scanners with reduced radiation dose.

\bibliographystyle{splncs04}
\bibliography{reference}

\newpage
\setcounter{section}{0}
\setcounter{figure}{0}
\setcounter{table}{0}
\section{Supplementary Information}

\subsection{Configurations of the few-angle dedicated scanner}
\begin{figure}[htb!]
\centering
\includegraphics[width=1.00\textwidth]{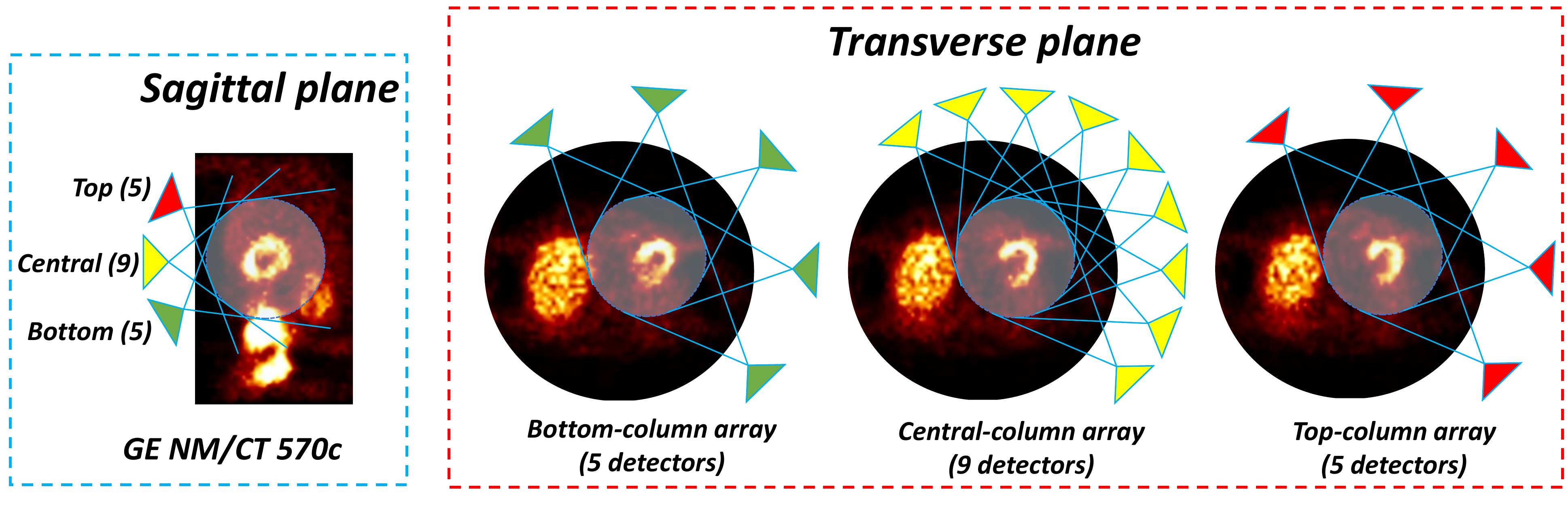}
\caption{(SI) Configurations of GE NM/CT 530c/570c and the latest few-angle scanners. GE 570c scanner comprises of 19 pinhole detectors arranged in three columns on a cylindrical surface (left blue box) with 5, 9, 5 detectors placed on bottom, central, and top columns, respectively (right red box). The most recent few-angle scanner only employ the 9 detectors at the central column for minimizing hardware expenses.}
\label{fig:configuration} 
\end{figure}

\subsection{Patient clinical characteristics in the dataset}

\begin{table} [htb!]
\caption{(SI) The gender, height, weight, and BMI distributions of the enrolled patients in the dataset. M stands for male, and F stands for female.}
\label{tab:data} 
\scriptsize
\centering
\begin{tabular}{ l | c | c | c | c | c}
\hline
\multicolumn{2}{l|}{\textbf{Datasets}}                          & \textbf{Age (year)}  & \textbf{Height (m)}  & \textbf{Weight (kg)}  & \textbf{BMI}      \Tstrut\Bstrut\\
\hline
\multirow{2}{*}{Training (108 M, 92 F)}  & Range                & $27 - 86$            & $1.32 - 2.03$        & $44.91 - 127.00$      & $18.10 - 48.05$   \Tstrut\Bstrut\\
                                         & Mean $\pm$ Std.      & $65.0 \pm 11.6$      & $1.68 \pm 0.11$      & $85.67 \pm 20.62$     & $30.29 \pm 6.57$  \Tstrut\Bstrut\\
\hline
\multirow{2}{*}{Validation (52 M, 22 F)} & Range                & $41 - 84$            & $1.47 - 1.85$        & $54.34 - 103.87$      & $19.53 - 38.11$   \Tstrut\Bstrut\\
                                         & Mean $\pm$ Std.      & $65.5 \pm 10.1$      & $1.70 \pm 0.09$      & $81.28 \pm 12.14$     & $28.33 \pm 4.43$  \Tstrut\Bstrut\\
\hline
\multirow{2}{*}{Testing (104 M, 96 F)}   & Range                & $39 - 87$            & $1.47 - 1.98$        & $45.00 - 140.00$      & $18.26 - 48.44$   \Tstrut\Bstrut\\
                                         & Mean $\pm$ Std.      & $64.2 \pm 10.7$      & $1.69 \pm 0.11$      & $86.41 \pm 18.91$     & $30.28 \pm 5.81$  \Tstrut\Bstrut\\
\hline  
\end{tabular}
\end{table}

\newpage
\subsection{Segment-wise evaluations of the Reconstructed SPECT}
\begin{figure}[htb!]
\centering
\includegraphics[width=0.90\textwidth]{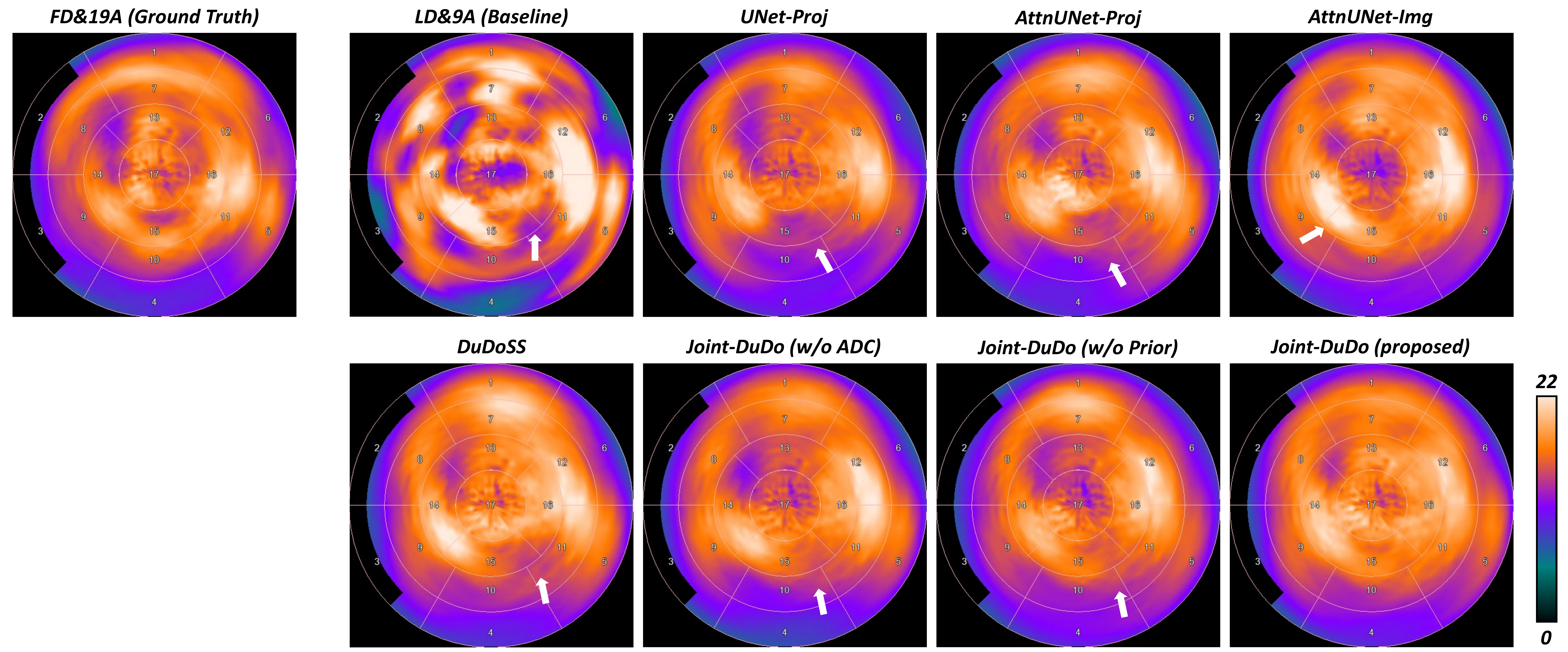}
\caption{(SI) Standard 17-segment polar maps of FD$\&$19A SPECT images \textbf{without} CT-based attenuation correction. White arrows denote the image inconsistencies. It can be observed that CDI-Net generated the most accurate cardiac polar maps. The single- or dual-domain groups as well as the ablation study groups show under- or over-estimation of the segment intensities.}
\label{fig:polarmapnc}
\end{figure}

\begin{figure}[htb!]
\centering
\includegraphics[width=0.90\textwidth]{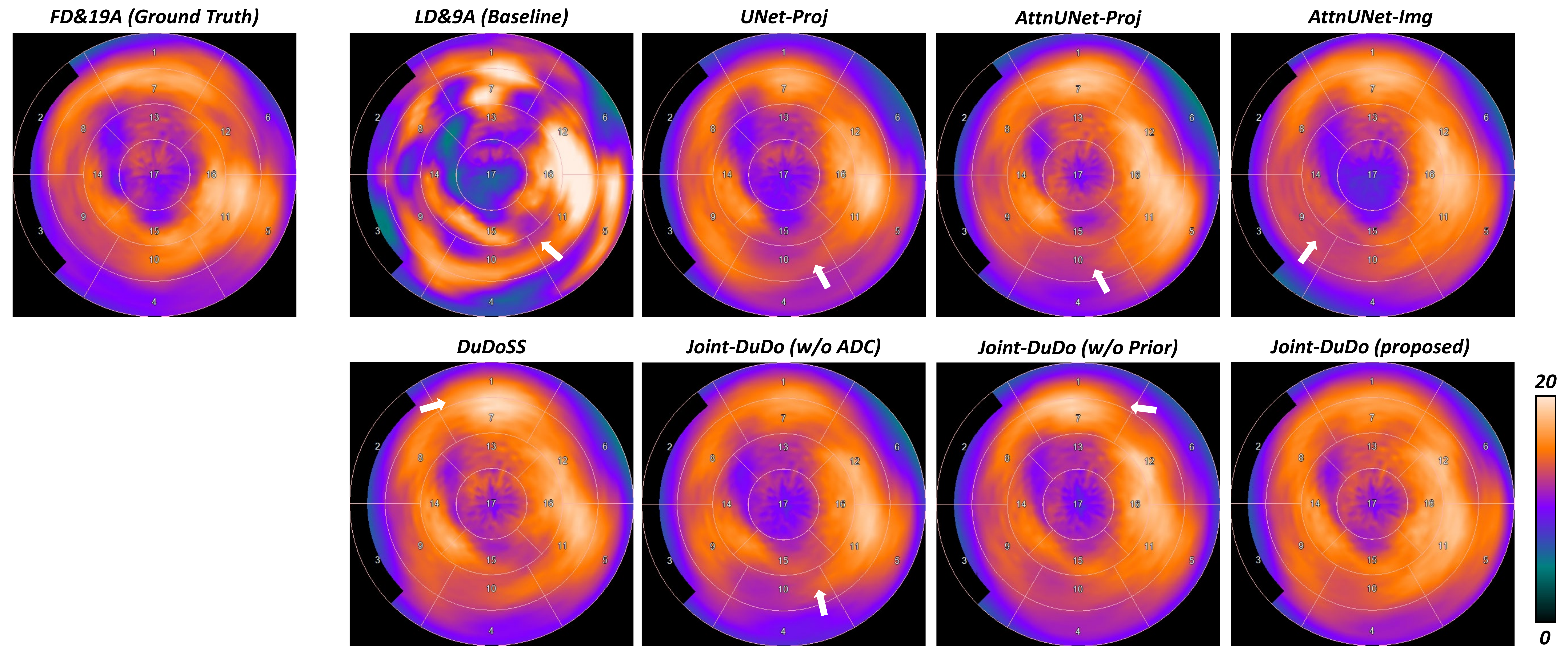}
\caption{(SI) Standard 17-segment polar maps of FD$\&$19A SPECT images \textbf{with} CT-based attenuation correction. White arrows point out the image inconsistencies. We can observe that CDI-Net generated the most accurate cardiac polar maps. The single- or dual-domain groups as well as the ablation study groups show under- or over-estimation of the segment intensities.}
\label{fig:polarmapac}
\end{figure}

\end{document}